\documentclass[journal]{IEEEtran}
\usepackage{times}  
\usepackage{helvet} 
\usepackage{courier} 
\usepackage{multirow}
\usepackage{array}
\usepackage{amsmath}
\usepackage{amssymb} 
\usepackage{caption}
\usepackage{subcaption}
\usepackage{float}
\usepackage{booktabs}
\usepackage[hyphens]{url}  
\usepackage{graphicx} 
\usepackage[numbers, square]{natbib}
\usepackage{caption} 
\usepackage{algorithm}
\usepackage{algorithmic}
\usepackage{newfloat}
\usepackage{listings}
\usepackage{amsmath}
\usepackage{hyperref}

\hypersetup{colorlinks=true,hypertex=true,linkcolor=blue,anchorcolor=red,citecolor=green}
\newcommand{\ie}{i.e.}
\ifCLASSINFOpdf
\else
\fi
\begin{document}

%
\title{PVAFN: Point-Voxel Attention Fusion Network with Multi-Pooling Enhancing\\ for 3D Object Detection}
%
%
%

\author{  
Yidi Li\textsuperscript{1,2}\quad 
Jiahao Wen\textsuperscript{1}\quad 
Bin Ren\textsuperscript{3,4}\quad
Wenhao Li\textsuperscript{2}\quad 
Zhenhuan Xu\textsuperscript{1}\quad
\thanks{$\star$Corresponding author: Hao Guo, guohao@tyut.edu.cn}$^{\star}$Hao Guo\textsuperscript{1}\quad
Hong~Liu\textsuperscript{2}\quad
Nicu Sebe\textsuperscript{4}\\
\textsuperscript{1}College of Computer Science and Technology, Taiyuan University of Technology\\
\textsuperscript{2}Key Laboratory of Machine Perception, Peking University \\ 
\textsuperscript{3}University of Pisa\quad 
\textsuperscript{4}University of Trento  
}

\markboth{}%
{}

\maketitle
\begin{abstract}
The integration of point and voxel representations is becoming more common in LiDAR-based 3D object detection. However, this combination often struggles with capturing semantic information effectively. Moreover, relying solely on point features within regions of interest can lead to information loss and limitations in local feature representation.
To tackle these challenges, we propose a novel two-stage 3D object detector, called Point-Voxel Attention Fusion Network (PVAFN). PVAFN leverages an attention mechanism to improve multi-modal feature fusion during the feature extraction phase. In the refinement stage, it utilizes a multi-pooling strategy to integrate both multi-scale and region-specific information effectively. The point-voxel attention mechanism adaptively combines point cloud and voxel-based Bird's-Eye-View (BEV) features, resulting in richer object representations that help to reduce false detections. Additionally, a multi-pooling enhancement module is introduced to boost the model's perception capabilities. This module employs cluster pooling and pyramid pooling techniques to efficiently capture key geometric details and fine-grained shape structures, thereby enhancing the integration of local and global features.
Extensive experiments on the KITTI and Waymo datasets demonstrate that the proposed PVAFN achieves competitive performance. The code and models will be available.
\end{abstract}


\IEEEpeerreviewmaketitle

\section{Introduction}
\label{sec:introduction}
Recently, LiDAR point cloud-based 3D object detection has become a focal point for addressing essential perception tasks in autonomous driving. LiDAR sensors are valued for their ability to provide precise distance measurements in diverse conditions, making them widely applicable in 3D tasks such as odometry, mapping \cite{ref1,ref3,ref2}, object tracking \cite{ref4,ref6,ref5}, and detection \cite{ref7,ref8,ref9}. However, the raw point cloud data is inherently unordered, sparse, and irregular, which complicates feature extraction and limits the effectiveness of traditional image-based methods. Consequently, it is imperative to develop and refine techniques specifically tailored for the analysis of point cloud data.

\begin{figure}[ht]
    \centering
    \includegraphics[width=0.9\columnwidth]{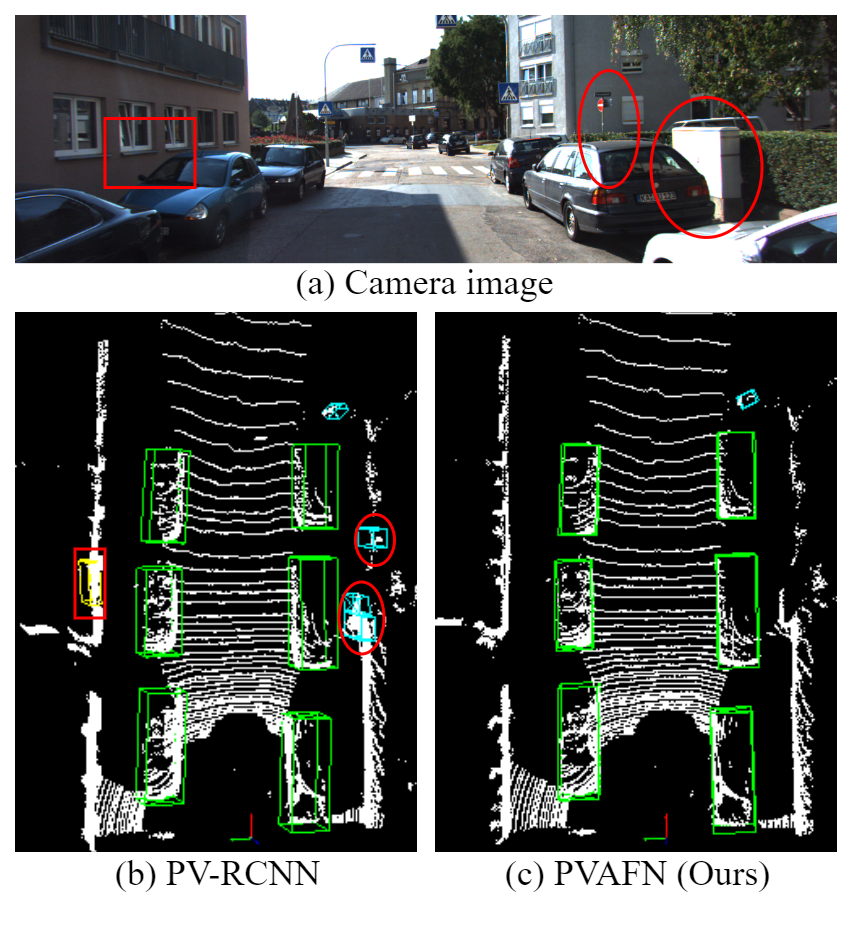}
    \vspace{-0.5cm}
    \caption{Comparison between (b) PV-RCNN~\cite{ref8} and (c) our PVAFN. Green, blue, and yellow boxes denote predicted cars, pedestrians, and cyclists, while red boxes are false detections. PVAFN effectively minimizes false detections, such as utility boxes, road signs, and walls.}
    \label{fig1}
    \vspace{-0.5cm}
\end{figure}
Based on the data type, current LiDAR-based 3D object detection methods can be broadly divided into two categories, \ie, the voxel-based and point-based methods. 
Voxel-based methods convert irregular point clouds into regular voxels and then learn high-dimensional features through 3D convolution \cite{ref10,ref7,ref11,ref50,ref51}. 
Usually, it is hard for voxel-based methods to find a balance between accuracy and efficiency, \ie, smaller voxels bring higher accuracy at the cost of higher computational cost, while using more significant voxels may ignore potential local details within the voxels. 
In contrast, point-based methods directly take point clouds as input and extract keypoint features with PointNet \cite{ref12} or its variants \cite{ref14,ref15,ref13}. 
Although this approach minimizes information loss compared to voxelization, efficiently capturing the local geometric structure and fine-grained details of the point cloud remains challenging due to the data's inherent sparsity and irregularity.


An alternative approach is to combine point-based and voxel-based representations, which can strike a balance between capturing richer geometric information and reducing computational cost. 
This has been validated by~\cite{ref8}, though it still faces limitations in effectively representing points. Specifically, the naive concatenation operation may yield suboptimal results in inferring hierarchical spatial relationships and semantic contexts within the point cloud. Moreover, inadequate point sampling within a Region of Interest (RoI) during the refinement stage often leads to inaccuracies in determining object size and category. This issue is illustrated in Fig.~\ref{fig1}, where Fig.~\ref{fig1} (a) shows the input camera image, Fig.~\ref{fig1} (b) displays the detection results from PV-RCNN\cite{ref8}, and Fig.~\ref{fig1} (c) presents the more accurate detections achieved by the proposed Point-Voxel Attention Fusion Network (PVAFN).


To address the challenge of insufficient semantic feature extraction when combining points and voxels, this paper introduces a novel two-stage 3D object detector with a point-voxel attention fusion module. This module involves three key steps: first, it integrates voxel features with BEV features to create multi-dimensional fusion features, differing from JPV-Net \cite{ref16} by recognizing both voxels and BEV as regular point cloud representations in distinct dimensions. Next, the combined keypoint and hybrid features are processed through a self-attention layer \cite{ref17} to enhance contextual information. Finally, the point-voxel attention fusion module adaptively merges point-wise features with voxel-BEV fusion features to improve semantic feature extraction.


Moreover, to address challenges such as point cloud sparsity, information loss, and limitations in local feature extraction, we propose a multi-pooling enhancement module for the refinement stage, which includes a RoI clustering pooling head and a RoI pyramid pooling head. The RoI clustering pooling head uses a density-based method to pinpoint key feature positions and aggregate features around density center points, thereby enhancing the capture of geometric information and eliminating background noise. The RoI pyramid pooling head incrementally expands the RoI and generates uniform grid points at each layer to form a grid pyramid \cite{ref18}, thereby improving global structure understanding through contextual information. The combined use of these heads significantly enhances detection accuracy.

Our main contributions are summarized as follows:
\begin{itemize}
    \item We introduce a novel Point-Voxel Attention Fusion Network (PVAFN) for 3D object detection. PVAFN enhances feature representation by adaptively integrating point features with voxel-BEV fusion features through a module that combines self-attention and point-voxel attention, enriching contextual information.
    \item We propose a multi-pooling enhancement module that combines the RoI clustering pooling head and the RoI pyramid pooling head to efficiently capture key geometric details and fine-grained shapes, thereby enhancing local and global perceptions.
    \item Extensive experiments on the KITTI and Waymo 3D object detection datasets validate the effectiveness of PVAFN, demonstrating competitive performance in detecting cars, pedestrians, and cyclists.
\end{itemize}
\section{Related Work}
\label{sec:related-work}

\subsection{Single Representation-Based 3D Object Detection.}
Single representation-based 3D object detection includes voxel-based~\cite{ref21,ref9} and point-based~\cite{ref15,ref25} methods, each with unique benefits and challenges. Voxel-based methods like VoxelNet~\cite{ref10} use voxel grids and 3D CNNs for feature extraction, but face high computational costs, partially reduced by SECOND~\cite{ref7} with sparse CNNs. Advances also include BEV maps and pillar-based representations that leverage 2D CNNs for real-time detection~\cite{ref19,ref20}. Despite improvements, voxel-based methods still contend with quantization errors. In contrast, point-based methods like PointNet~\cite{ref12} process point clouds directly without grid projection, using permutation-invariant layers to extract features. PointRCNN~\cite{ref22} and subsequent works~\cite{ref23} refine PointNet with two-stage frameworks and hybrid sampling. These methods offer enhanced flexibility and geometric details, often surpassing voxel-based techniques in detection performance. Both voxel-based and point-based approaches have advanced the field, each with unique benefits and limitations.

\begin{figure*}[t]
    \centering
    \includegraphics[width=1.0\textwidth]{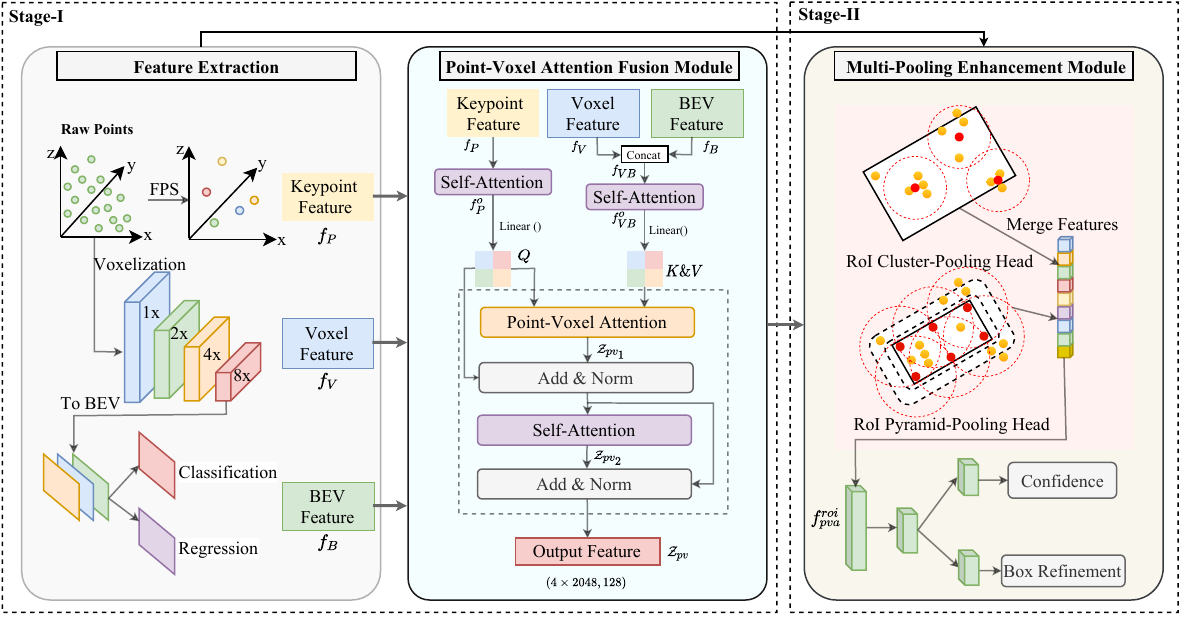}
    \vspace{-1em}
    \caption{Overall architecture of the proposed PVAFN. First, the raw point cloud undergoes keypoint sampling and voxelization. The resulting keypoint, voxel, and BEV features are fused using the point-voxel attention fusion module, which employs self-attention, point-voxel attention, and residual connections. The multi-pooling enhancement module then extracts geometric and fine-grained features for proposal generation and refinement.}
    \label{fig2}
\end{figure*}

\subsection{Point-Voxel Representation-Based 3D Object Detection.}
Recently, methods have emerged that integrate the benefits of both point-based and voxel-based representations~\cite{ref8,ref16}. For instance, STD~\cite{ref23} uses PointNet++~\cite{ref14} for initial extraction and point pooling to refine features into voxel representations. Similarly, SA-SSD~\cite{ref27} introduces an auxiliary network that supervises voxel context to improve focus on intra-object structures. PV-RCNN~\cite{ref8} uses sparse convolution to generate 3D proposals and a voxel set abstraction module to refine features through keypoint extraction. The proposed PVAFN improves integration by preserving detailed point features and combining them with voxel-BEV hybrid features via the attention mechanism, achieving superior performance in merging point and voxel information than others.

\subsection{Vision Transformer (ViTs) for Point Cloud Analysis.}
Transformers~\cite{ref17,ref49}, originally for NLP, now excel in computer vision and 3D point clouds~\cite{khan2022transformers,PointBERT,ma2024implicit,ren2024bringing}. Pointformer~\cite{ref13} uses local-global multi-head self-attention (MSA) for 3D point clouds, while Group-Free~\cite{ref28} applies MSA to all points, removing the need for handcrafted grouping. 3DETR~\cite{ref29} provides an end-to-end ViTs-based model with minimal 3D bias. VoTr~\cite{ref30} uses voxel-based ViTs with local and dilated attention to expand the receptive field. SST~\cite{ref31} applies MSA to non-empty voxels in shifted 3D windows. DGCNN~\cite{ref32} integrates BEV features with deformable attention, and VISTA~\cite{ref33} fuses global multi-view features via MSA. Despite advances, these methods often lose fine-grained details due to voxelization. PVAFN addresses this by adaptively combining point-wise features with voxel-BEV fusion, capturing richer contextual information.
\section{The Proposed Method}
\label{sec:method}
In this paper, we propose PVAFN, a novel two-stage 3D object detection network, detailed in Fig.~\ref{fig2}. In stage I, downsampling and voxelization methods, similar to PV-RCNN~\cite{ref8}, are used to obtain keypoint features $f_P$, voxel features $f_V$, and BEV features $f_B$ with the feature extraction module (left part of Stage-I in Fig.~\ref{fig2}). These features are then processed by the proposed point-voxel attention fusion module to enhance contextual representation. In stage II, the multi-pooling enhancement module, comprising the RoI clustering pooling head for key geometric information and the RoI pyramid pooling head for fine-grained shape feature extraction, refines these features for classification and regression. Details of these components are provided in the following subsections.

\subsection{Point-Voxel Attention Fusion Module}
\noindent\textbf{Motivation.}
Existing methods have demonstrated that ViTs can improve 3D object detection accuracy~\cite{ref13,ref30,ref29,ref31}. 
For example, VoTr~\cite{ref30} innovatively applies MSA to both empty and non-empty voxel locations using sparse and submanifold voxel modules, establishing long-range relationships between voxels through an efficient attention operation. Despite these advances, voxel-only methods with ViTs still face information loss issues.
Based on this observation, we conclude that a promising solution should make use of different kinds of information as much as possible. To this end, we propose the Point-Voxel Attention Fusion Module, which is designed to well integrate three kinds of 3D features with attention operation.



\subsubsection{Main Pipeline.} 
\begin{figure}[t]
    \centering
    \includegraphics[width=1.0\columnwidth]{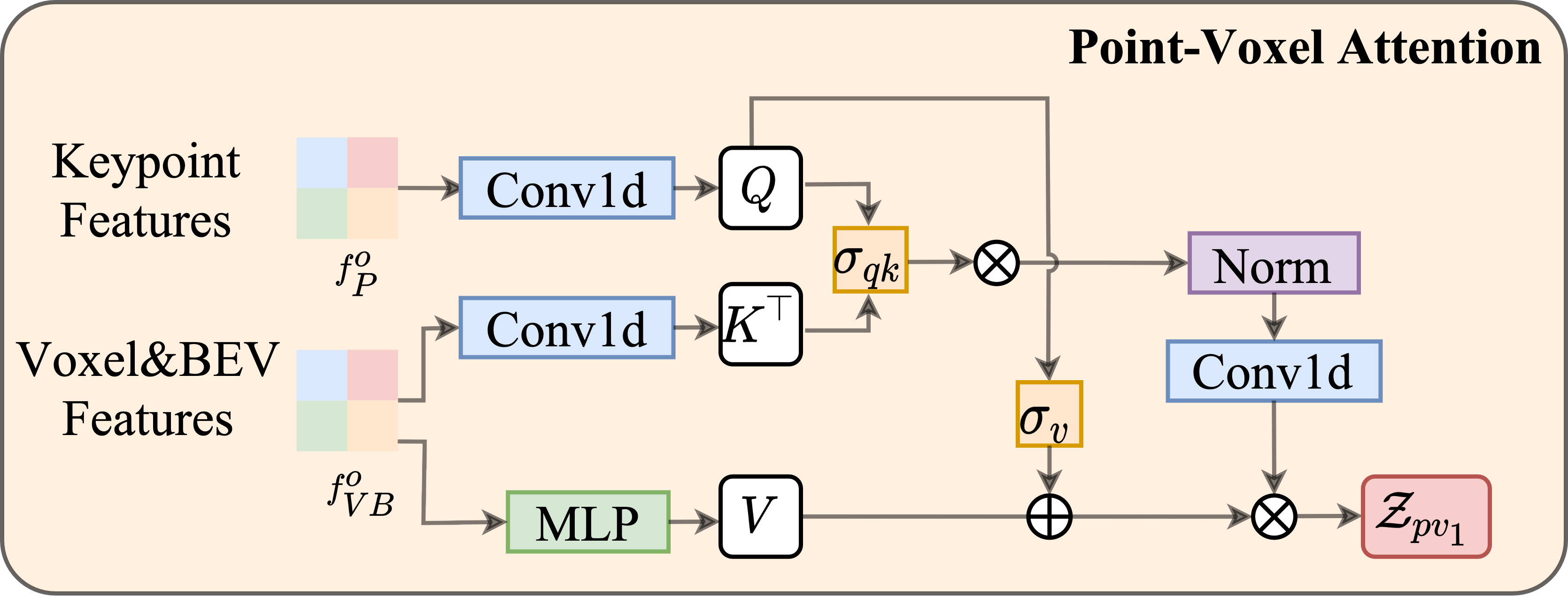}
    \vspace{-1.3em}
    \caption{Illustration of Point-Voxel Attention. It introduces a learnable gating function $\sigma_*$, composed of attention and graph operators, which can dynamically select attention components to achieve different optimization effects.}
    \label{fig3}
    \vspace{-1.2em}
\end{figure}

As shown in Fig.~\ref{fig2}, given the keypoint features $f_P$, voxel features $f_V$, and BEV features $f_B$ from the feature extraction module of Stage-I. $f_P$ first go through a standard self-attention operation for keypoint feature aggregation and outputs $f_{P}^{o}$. Meanwhile, $f_V$ and $f_B$ are combined via concatenation operation for achieving better geometric advantages based on the inborn attributes from both the voxel and BEV features, forming $f_{VB}$. 
Similarly, another self-attention operation is utilized for $f_{VB}$ and outputs $f_{VB}^{o}$. 
Then $f_{P}^{o}$ and $f_{VB}^{o}$ are linearly project to $Q$ and $K \& V$ (See Fig.~\ref{fig3}). We formulate this process as $Q = \mathbf{W_1}f_{P}^{o}$, $K = \mathbf{W_2}f_{VB}^{o}$, and $V = \mathbf{W}_{MLP}f_{VB}^{o}$, respectively. $\mathbf{W_1}$ and $\mathbf{W_2}$ are the learnable weight of 1D CNN while $\mathbf{W}_{MLP}$ indicates the learnable weight of the MLP operation. Finally, $Q$, $K$, and $V$ pass through the proposed point-voxel attention for better feature fusion.  

\subsubsection{Premilinaries.}
Usually, the standard attention that commonly adopted for 3D point cloud~\cite{ref13} is formalized as follows:
\begin{equation}
    \mathcal{Z} = \sum_{i \in \Omega(r)} \operatorname{Softmax}(Q_i K_i^{\top}) \odot V_i,
\end{equation}
where $\operatorname{Softmax}(\cdot)$ denots the Softmax function. $\mathcal{Z}$ is the fused feature. Then, \cite{ref36} proposed the point attention operators which works as an extension of the standard attention operator-based method, focusing more on the relationship between the position and features of points, and is defined as:
\begin{equation}
    \mathcal{Z}_{pts} = \sum_{i \in \Omega(r)} \operatorname{Softmax}(Q_i + K_i) \odot (V_i + Q_i).
\end{equation}

\subsubsection{Point-Voxel Attention.} 
The point attention operator-based method overly emphasizes the relationship between the position and features of points. However, experiments have shown that it is unsuitable for the relationship between keypoints and voxel features. Inspired by the above two approaches, we propose point-voxel attention, which fully considers the structural similarity and contextual relationships corresponding to keypoints features and voxel-BEV fusion features, with the following formulation:
\begin{equation}
    \mathcal{Z}_{pv_1} = \sum_{i \in \Omega(r)} \operatorname{Softmax}(\sigma_{qk} Q_i K_i^{\top}) \odot (V_i + \sigma_v Q_i),
    \label{eq:pvatt}
\end{equation}
where $\sigma_{qk}$ and $\sigma_v$ are learnable gating functions implemented by linear projections of the output of the embedding by the sigmoid activation function. Through the gating functions $\sigma_{qk}$ and $\sigma_v$, the point-voxel attention can selectively learn key features from the point-pixel features and balance the proportion of features used for fusion, respectively. Note that we omit the norm and 1D CNN in Eq.~\ref{eq:pvatt}. $\mathcal{Z}_{pv_1}$ now contains the well-fused feature with attributes from both points and voxel. To further well-aggregate both information, another self-attention is applied to $\mathcal{Z}_{pv_1}$ and outputs $\mathcal{Z}_{pv_2}$, and then a skip-connection and a normalization is applied to $\mathcal{Z}_{pv_2}$, forming our final fused feature $\mathcal{Z}_{pv}$. 


\subsection{Multi-Pooling Enhancement Module}
\begin{figure}[t]
    \centering
    \includegraphics[width=1.0\columnwidth]{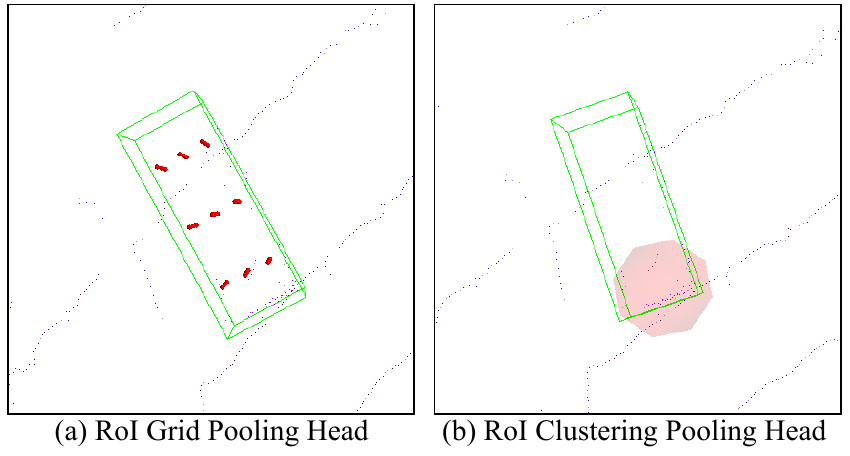}
    \vspace{-1.8em}
    \caption{Comparison of cluster pooling head and grid pooling head. For sparse and uneven RoIs, (a) (\ie, the grid pooling method) fuses all points within the RoI, including background information such as the ground. In contrast, (b) (\ie, the clustering pooling method) focuses only on the key geometric information of the target within the bounding box.}
    \label{fig4}
    \vspace{-1.5em}
\end{figure}

During the refinement process, we observe that some RoIs contain sparse points and extremely incomplete object shapes. As shown in Fig.~\ref{fig4}, keypoints within the RoI are sparsely concentrated in a specific location, and there are background points, such as ground information within the RoI. 
Fig.~\ref{fig4}(a) shows that using only the grid pooling head is insufficient to infer object classes, while Fig.~\ref{fig4}(b) indicates that the cluster pooling head can effectively address this issue.
To accurately infer geometric information and object categories, the precise location of key features and sufficient neighboring point data are essential. Therefore, we propose the multi-pooling enhancement module for RoI feature extraction, which includes a RoI clustering pooling head and a RoI pyramid pooling head.

\subsubsection{RoI Clustering Pooling Head.} 
In previous refinement network studies, most feature extraction was conducted using RoI grid structures based on spherical radius search~\cite{ref8}. Although RoI grid pooling can obtain global RoI feature information, it is not easy to focus on extracting key foreground point features for some sparse and unevenly distributed points. To address this, we propose a RoI Clustering Pooling Head based on DBSCAN (Density-Based Spatial Clustering of Applications with Noise), which compensates for the shortcomings of RoI grid pooling, quickly locates key features, captures key geometric information, and removes background noise. For each 3D candidate box $b_i = (x_i, y_i, z_i, h_i, w_i, l_i, \theta_i)$, the RoI Clustering Pooling Head slightly enlarges its dimensions to create a new 3D box $b'_i = (x_i, y_i, z_i, h_i + \varphi, w_i + \varphi, l_i + \varphi, \theta_i)$, where $\varphi$ is a constant used to expand the bounding box size. Interest points within the bounding box are clustered using DBSCAN, with clustering based on a specified search radius and minimum cluster size. For feature extraction, the process begins by calculating the average position of all points within each cluster. Subsequently, the features of all points in the cluster are aggregated around this average point.

\subsubsection{RoI Pyramid Pooling Head.}
The RoI pyramid pooling head extracts RoI features by constructing a pyramid grid structure. First, the original RoI region size is gradually expanded, and then uniform grid points are generated within the RoI at each level to construct the grid pyramid \cite{ref18}. The RoI pyramid pooling head contains grid points inside and outside the RoI. The grid points inside the RoI can capture fine-grained shape structures for precise box refinement, and the grid points outside the RoI can capture extensive global contextual information for recognizing incomplete objects. The coordinates of the grid points are determined by the expanded RoI and grid sizes, expressed as:
\begin{equation}
   p_{grid} = \left(\frac{\rho_l l_i}{n_l}, \frac{\rho_w w_i}{n_w}, \frac{\rho_h h_i}{n_h}\right) \cdot (0.5 + (i,j,k)) + (x_i, y_i, z_i),
\end{equation}
where $l_i$, $w_i$, and $h_i$ represent the length, width, and height of the candidate box, $(x_i, y_i, z_i)$ is the coordinate of the candidate box, $(n_l, n_w, n_h)$ represents the grid size, and $\rho_l$, $\rho_w$, and $\rho_h$ represent the RoI size control rates. The grid points of each RoI level are grouped as sphere centers, and features are extracted through point-voxel attention. Let $p_i, f_i$ be the coordinates of the $i$-th interest point near $p_{grid}$ and its corresponding feature vector. The process is as follows:
\begin{equation}
    f_{pvAtt}^{roi} = \sum_{i \in \Omega(r)} \operatorname{Softmax}(\sigma_{qk} Q^{pos}_i K_i^{\top}) \odot (V_i + \sigma_v Q^{pos}_i),
\end{equation}
where $Q_{pos}^i$ is the query embedding from $p_{grid}$ to $p_i$, $K^i = \operatorname{Linear}(f_i)$, $V^i$ is the value embedding obtained from $f_i$, and $\sigma_{qk}$ and $\sigma_v$ are learnable gating functions. The point-voxel attention can effectively aggregate surrounding point features and preserve the internal structure. Finally, the RoI features from different levels are stacked and fused to obtain the grid point features of the RoI pyramid pooling head, which are then fused with the features from the RoI clustering pooling head to obtain enhanced features for refined localization and regression.

\subsection{Optimization Objectives}
The PVAFN framework is trained end-to-end, and the overall loss function consists of the first-stage RPN loss and the second-stage box refinement loss. For 3D proposal generation, we follow SECOND \cite{ref7} and design a region proposal loss $\mathcal{L}_{rpn}$, which uses focal loss with default hyperparameters for classification, smooth-L1 loss for box regression, and cross-entropy loss for orientation, as follows:
\begin{equation}
\mathcal{L}_{rpn} = \mathcal{L}_{cls}^P + \beta \mathcal{L}_{reg}^P,
\end{equation}
\begin{equation}
\mathcal{L}_{cls}^{rpn} = -\frac{1}{N} \sum_{i=1}^N \alpha (1 - p_i)^\gamma \log \mu_i, \\  
\end{equation}
\begin{equation}
\mathcal{L}_{reg}^{rpn} = -\frac{1}{N} \sum_{r \in x,y,z,l,h,w,\theta}^N \mathcal{L}_s (\Delta \hat{r}, \Delta r),
\end{equation}
where $\mathcal{L}_{cls}^P$ and $\mathcal{L}_{reg}^P$ represent the classification and regression losses, $\alpha$ and $\gamma$ are prediction hyperparameters, $\mu_i$ is the probability of predicting the foreground point, $\Delta \hat{r}$ represents the predicted residual of the candidate box, and $\mathcal{L}_s$ represents smooth-L1 loss. 

Similarly, the second-stage box refinement loss $\mathcal{L}_{rcnn}$ includes classification and regression loss functions. The regression loss function is defined the same as in the RPN, while the classification loss function is defined as:
\begin{equation}
    \mathcal{L}_{cls}^{rcnn} = -\frac{1}{N} \sum_{i=1}^C \left( y_i \log \hat{y}_i + (1 - y_i) \log (1 - \hat{y}_i) \right),
\end{equation}
where $\mathcal{L}_{cls}^{rcnn}$ is the classification loss, $y_i$ is the target classification label, $\hat{y}_i$ is the target prediction probability, and $N$ is the number of targets. Notably, in typical 3D object detection algorithms, when setting classification labels based on the IoU value, the classification label is set to $1$ when the IoU $\sigma_{iou}$ is greater than the foreground threshold $\sigma_{fg} = 0.75$, set to 0 when it is less than the background threshold $\sigma_{bg} = 0.25$, and set to $-1$ when $\sigma_{iou}$ is greater than the background threshold but less than the foreground threshold. 

However, this approach needs to pay more attention to key information within the target box. To fully learn all features, we calculate the average classification loss for all points and set the target classification label as:
\begin{equation}
    y_i = \frac{\sigma_{iou} - \sigma_{bg}}{\sigma_{fg} - \sigma_{bg}}, \quad y_i \in (0, 1),
\end{equation}
where $\sigma_{iou}$ is the IoU value, $\sigma_{bg}$ is the background threshold, and $\sigma_{fg}$ is the foreground threshold.
\section{Experiments and Discussions}
\label{sec:experiments}
\subsection{Implementation Details}
\subsubsection{Datasets.} 
We evaluate the proposed model on the KITTI~\cite{ref37} and Waymo~\cite{ref38} datasets. The KITTI dataset is a widely used autonomous driving benchmark containing 7,481 training samples and 7,518 test samples with three categories, \ie, cars, pedestrians, and cyclists. The training samples are split into a training set (3,712 samples) and a validation set (3,769 samples). Average Precision (AP) is used to evaluate all three difficulty levels (\ie, easy, moderate, and hard). Our model is trained on the training set and evaluated on the validation set. 
The Waymo dataset consists of 798 scenes for training and 202 scenes for validation. The evaluation protocol consists of AP and Average Precision weighted by Heading (APH). Moreover, it includes two difficulty levels: Level 1 indicates objects containing more than 5 points, while Level 2 indicates objects containing at least 1 point.
\begin{table*}[!ht]
    \centering
    \setlength\tabcolsep{7.5pt}
    \footnotesize
    \begin{tabular}{l|l|ccc|ccc|ccc}
    \toprule[0.95pt]
        \multirow{2}{*}{Types} & \multirow{2}{*}{Methods} & \multicolumn{3}{c|}{Car 3D AP\textsubscript{R40}} & \multicolumn{3}{c|}{Pedestrian 3D AP\textsubscript{R40} } & \multicolumn{3}{c}{Cyclist 3D AP\textsubscript{R40}} \\ 
        ~ & ~ & Easy & Mod. & Hard & Easy & Mod. & Hard & Easy & Mod. & Hard \\ \midrule[0.6pt]
        \multirow{5}{*}{1-Stage} & SECOND (Yan et al. 2018)& 87.12 & 79.3 & 75.91  & 50.66 & 47.82 & 40.54 & 80.31 & 64.98 & 61.01 \\ 
        ~ & PointPillars \cite{ref20} & 84.01 & 76.11 & 72.19  & 59.45 & 50.81 & 44.98 & 85.10 & 65.65 & 60.32  \\ 
        ~ & CIA-SSD \cite{ref39} & 90.57 & 81.33 & 78.85  & - & - & - & - & - & -  \\ 
        ~ & SVGA-Net \cite{ref40} & 88.93 & 81.87 & 79.13  & 56.05 & 50.44 & 43.93 & 86.16 & 69.08 & 62.96  \\ 
        ~ & IA-SSD \cite{ref41} & 90.47 & 81.72 & 78.20  & 55.90 & 50.03 & 44.00 & \underline{90.23} & 73.25 & 68.41  \\ \midrule[0.6pt]
        \multirow{7}{*}{2-Stage} & PointRCNN (Shi et al. 2019) & 88.96 & 77.05 & 74.50  & 56.11 & 48.41 & 42.33 & 83.54 & 65.82 & 60.33  \\ 
        ~ & Part-A\textsuperscript{2} \cite{ref21} & 90.23 & 80.45 & 77.65  & 58.77 & 51.89 & 46.25 & 85.40 & 68.82 & 64.55  \\ 
        ~ & Pyramid-PV \cite{ref18} & 90.00 & 83.48 & 80.09  & - & - & - & - & - & -  \\ 
        ~ & VoTr-TSD \cite{ref30} & 91.30 & 83.42 & \underline{81.44}  & - & - & - & - & - & -  \\ 
        ~ & PG-RCNN \cite{ref26} & 90.97 & \underline{83.53} & 80.83  & 56.79 & 50.04 & 44.69 & \textbf{90.47} & \textbf{74.12} & \textbf{69.82}  \\ 
        ~ & PV-RCNN \cite{ref8} & \underline{91.86} & 82.85 & 80.31  & \underline{59.97} & \underline{52.37} & \underline{46.59} & 86.89 & 70.65 & 66.36  \\ 
        ~ & \textbf{PVAFN (Ours)} & \textbf{92.81} & \textbf{83.92} & \textbf{81.92}  & \textbf{61.98} & \textbf{54.01} & \textbf{48.94} & 88.50 & \underline{73.54} & \underline{68.93}  \\ 
        \bottomrule[0.95pt]
    \end{tabular}
    \vspace{-0.5em}
    \caption{Comparison results with State-of-The-Art methods on the KITTI validation set. Best performance values are shown in bold, and second-best performance values are underlined.}
    \label{table1}
    \vspace{-1em}
\end{table*}

\begin{table}[t]
    \centering
    \footnotesize
    \begin{tabular}{l|ccc}
    \toprule[0.95pt]
        \multirow{2}{*}{Methods} & \multicolumn{3}{c}{Car 3D AP\textsubscript{R40}}  \\ 
        ~ & Easy & Mod. & Hard  \\  \midrule[0.6pt]
        SECOND (Yan et al. 2018) & 82.02 & 72.68 & 66.27  \\
        CIA-SSD \cite{ref39} & \underline{89.59} & 80.28 & 72.87  \\ 
        SVGA-Net \cite{ref40} & 87.33 & 80.47 & 75.91  \\ 
        IA-SSD \cite{ref41} & 88.87 & 80.32 & 75.10  \\ 
        PointRCNN (Shi et al. 2019) & 86.96 & 75.64 & 70.70  \\ 
        Part-A\textsuperscript{2} \cite{ref21} & 87.81 & 78.49 & 73.51  \\ 
        GD-MAE \cite{ref42} & 88.14 & 79.03 & 73.55  \\ 
        P-PV-RCNN++ (Chen et al. 2024) & 87.65 & 81.28 & 76.79  \\ 
        PV-RCNN \cite{ref8} & \textbf{90.25} & \underline{81.43} & \underline{76.82}  \\ 
        \textbf{PVAFN (Ours)} & 88.15 & \textbf{81.53} & \textbf{76.90}  \\ 
        \bottomrule[0.95pt]
    \end{tabular}
    \vspace{-0.5em}
    \caption{Performance comparison on KITTI test set.}
    \label{table2}
    \vspace{-1em}
\end{table}

\subsubsection{Network Architecture.}
For 3D scenes in the KITTI dataset, the detection range is set to [0, 70.4] along the X-axis, [-40, 40] along the Y-axis, and [-3, 1] along the Z-axis, containing approximately 20,000 LiDAR points. A voxel size of (0.05m, 0.05m, 0.1m) is used as the voxel input to voxelize each scene, and 2,048 points are sampled from the original point cloud as the point input. For the Waymo dataset, the detection range is set to (-75.2, 75.2), (-75.2, 75.2), and (-2, 4), with a voxel size of (0.1m, 0.1m, 0.15m). As shown in Fig.~\ref{fig2}, the voxel CNN consists of four 3D encoding levels and 2D convolutions for BEV maps, similar to the SECOND network. The feature dimension for the points is 32, the feature dimensions for the four voxel levels are (32, 64, 128, 128), and the feature dimension for the BEV map is 256. Then, the point feature dimensions are aligned with the voxel-BEV fused features through an MLP and passed to the point-voxel attention module, with the output of the 128-dimensional feature to the refinement network after multiple layers of attention. In the refinement network, the expansion range of the RoI in the RoI clustering pooling head is set to 0.4, with a minimum cluster size and radius set to 2 and 0.2, respectively. For the RoI pyramid pooling head, the official hyperparameters proposed by Pyramid-RCNN \cite{ref18} are followed.

\subsubsection{Training and Inference Schemes.}
The PVAFN framework is trained end-to-end using the ADAM optimizer, with an initial learning rate and weight decay set to 0.01 and a batch size of 1, running on 4 Tesla V100 GPUs. The learning rate is decayed using a cosine annealing strategy over 80 training epochs. During training, the IoU thresholds for positive and negative anchors are set to 0.7 and 0.25, respectively. For a fair comparison, other configurations are kept the same as the baseline network \cite{ref8}.

\subsection{Experimental Results}
In this section, we train the model on the training set and adjust hyperparameters based on the evaluation results on the validation set. We also submit the detection results for the Car class to the official KITTI detection server for evaluation. The server calculates performance on the test set using 40 recall positions. Additionally, we evaluate the model using the Waymo dataset and compare its performance with State-of-The-Art (SoTA) 3D object detection methods. The compared methods are highly representative or employ similar techniques to those used in this paper.

\begin{figure*}[t]
    \centering
    \includegraphics[width=2.2\columnwidth]{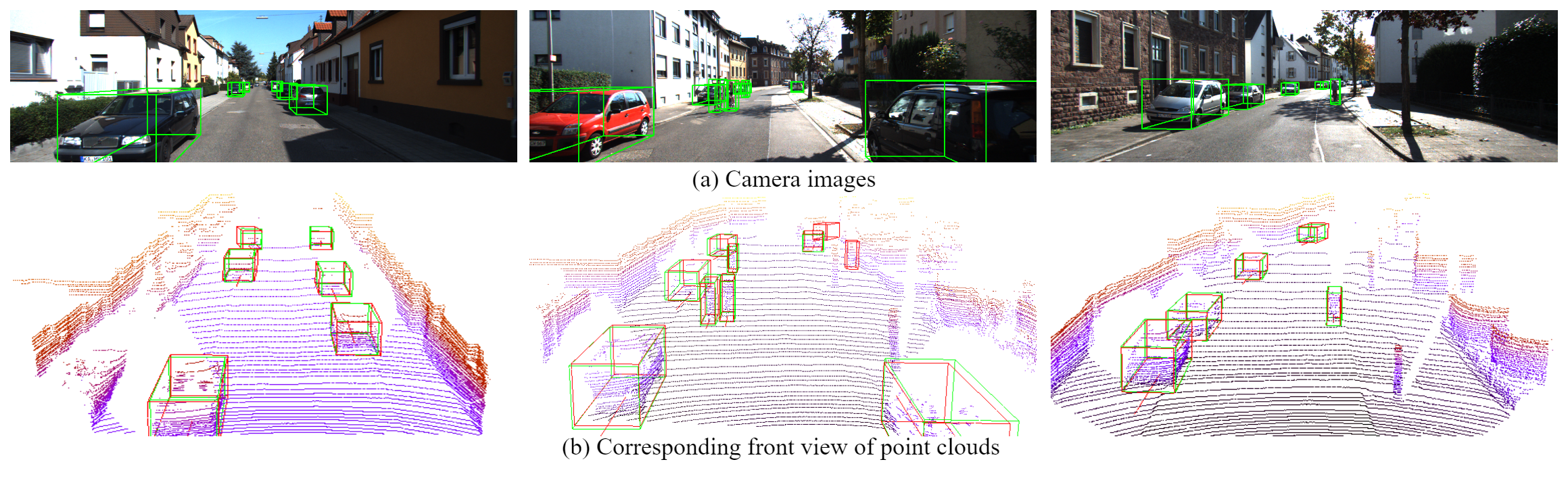}
    \vspace{-2.5em}
    \caption{Visualization results on the KITTI validation set. (a) shows the camera image, and (b) shows the corresponding front view of the point cloud. The ground truth bounding box is green, and the predicted bounding box is red.}
    \label{fig5}
    \vspace{-1em}
\end{figure*}

As shown in Tab.~\ref{table1}, in the 3D detection task, PVAFN significantly outperforms other state-of-the-art object detection models. Specifically, PVAFN improves by $0.95\%$, $1.07\%$, and $1.61\% $ over PV-RCNN in the Car category in the three difficulty levels. For the Pedestrian category, PVAFN improves by $2.01\%$, $1.64\%$, and $2.35\%$ over PV-RCNN in the three difficulty levels. For the Cyclist category, PVAFN improves by $1.61\%$, $0.89\%$, and $0.57\%$ over PV-RCNN in the three difficulty levels. Additionally, Fig.~\ref{fig5} shows the qualitative results of PVAFN on the KITTI validation set.

As shown in Tab.~\ref{table2}, we also compare PVAFN's performance on the KITTI test set in the Car category with the latest 3D object detection methods. Compared to PV-RCNN, PVAFN performs better in moderate and hard difficulty levels. In addition, Tab.~\ref{table3} shows the comparison results with other advanced methods on the Waymo validation set for the vehicle category, where PVAFN leads by $3.01\%–3.47\%$ at both levels due to its excellent ability to combine contextual information and capture key information.

\begin{table}[!t]
    \centering
    \setlength{\tabcolsep}{2.2pt} 
    \footnotesize
    \begin{tabular}{l|l|cc}
    \toprule[0.95pt]
        ~ & \multirow{2}{*}{Methods} & LEVEL\_1  & LEVEL\_2  \\ 
        ~ & ~ & 3D AP/APH & 3D AP/APH \\ 
        \midrule[0.6pt]
        \multirow{5}{*}{1} & SECOND (Yan et al. 2018) & 72.37/71.58 & 63.82/63.23 \\ 
        ~ & PointPillars \cite{ref20} & 71.56/70.87 & 63.15/62.59 \\ 
        ~ & CIA-SSD \cite{ref39} & 70.61/69.58 & 61.73/60.82 \\ 
        ~ & SWFormer \cite{ref44} & 77.80/77.30 & 69.20/68.80 \\ 
        ~ & IA-SSD \cite{ref41} & 70.51/69.63 & 61.55/60.82 \\ 
        \midrule[0.6pt]
        \multirow{7}{*}{2} & CT3D \cite{ref45} & 76.30/- & 69.04/- \\ 
        ~ & Part-A\textsuperscript{2} \cite{ref21} & 77.05/76.51 & 68.47/67.97 \\ 
        ~ & Pyramid-PV \cite{ref18} & 76.30/75.68 & 67.23/66.68 \\ 
        ~ & BtcDet (Xu et al. 2022) & \underline{78.58/78.06} & 70.10/69.61 \\ 
        ~ & PDV (Hu et al. 2022) & 76.85/76.33 & \underline{69.30/68.81} \\ 
        ~ & PV-RCNN \cite{ref8} & 77.51/76.89 & 68.98/68.41 \\ 
        ~ & \textbf{PVAFN (Ours)} & \textbf{80.93/80.36} & \textbf{72.01/71.61 }\\ 
        \bottomrule[0.95pt]
    \end{tabular}
    \vspace{-0.5em}
    \caption{Comparison on Waymo vehicle validation set. }
    \vspace{-1em}
    \label{table3}
\end{table}

\subsection{Ablation Experiments}
To validate the effectiveness of the proposed method, we conducted ablation studies on the KITTI validation set focusing on the Point-Voxel Attention Fusion Module and the Multi-Pooling Enhancement Module.

\subsubsection{Effects of Point-Voxel Attention Fusion Module.}
We compared our module against the Voxel Set Abstraction (VSA) module ~\cite{ref8} and other attention mechanisms to demonstrate the rationality of the design choices made for the Point-Voxel Attention Fusion Module components. 
Tab.~\ref{table4} shows the comparison results for 3D detection with 3 categories at the moderate difficulty level
First, we examined the performance gains brought by the MSA~\cite{ref17}. 
Next, we compared methods based on the Standard Attention Operator (SAO) \cite{ref13}, Point Attention Operator (PAO) \cite{ref36}, and Graph-based Attention Operator (GO) \cite{ref48}. Finally, we evaluated the effectiveness of using the Point-Voxel Attention alone and the combination of Self-Attention and Point-Voxel Attention. The results show that using the self-attention mechanism improves the AP (Average Precision) across all categories, highlighting the advantage of attention mechanisms in combining contextual information. Methods based on the Point Attention Operator and Graph-based Attention Operator are unsuitable for point-voxel level feature fusion. Our proposed Point-Voxel Attention outperforms other attention mechanisms, and its combination with Self-Attention significantly enhances detection performance across all categories.

\begin{table}[!t]
    \centering
    \footnotesize
    \setlength{\tabcolsep}{12pt} 
    \begin{tabular}{l|ccc}
    \toprule[0.95pt]
         \multirow{2}{*}{Point \& Voxel Fusion} &  \multicolumn{3}{c}{3D AP\textsubscript{R40} (Mod.)}   \\
        ~ & Car & Ped. & Cyc. \\ \midrule[0.6pt]
        VSA(PV-RCNN) & 82.85 & 52.37 & 70.65 \\ 
        SA & 83.14 & 52.42 & 70.81 \\ 
        SAO & 83.02 & 52.50 & 70.31 \\ 
        PAO & 82.00 & 52.40 & 68.44 \\ 
        GO & 82.18 & 52.31 & 70.14 \\ 
        PVA (Ours) & 83.54 & 53.01 & 70.89 \\ 
        SA+PVA (Ours) & \textbf{83.74} & \textbf{53.88} & \textbf{71.32} \\ 
        \bottomrule[0.95pt]
    \end{tabular}
    \vspace{-0.5em}
    \caption{Comparison of the proposed point-voxel attention fusion module with other methods.}
    \vspace{-1.5em}
    \label{table4}
\end{table}

\subsubsection{Effects of Multi-Pooling Enhancement Module.} 
The proposed multi-pooling enhancement module comprises a Clustering Pooling Head (CPH) and a Pyramid Pooling Head (PPH). 
We compared the performance of the RoI Grid Pooling Head (GPH)~\cite{ref8}, clustering pooling method, pyramid pooling method, and their combinations. In Tab.~\ref{table5}, PPH refers to the pyramid pooling head \cite{ref18}, and $\mathrm{PPH^{PVA}}$ refers to the pyramid pooling head improved by our attention mechanism. Experiments demonstrated that the RoI clustering pooling head outperforms the grid pooling head. The improved RoI pyramid pooling head slightly outperforms the original PPH. Notably, the multi-pooling enhancement module, which combines the clustering pooling head and pyramid pooling head, achieves superior performance, improving the average precision by $0.6\%-1.54\%$ across three categories compared to the grid pooling method.

\begin{table}[!t]
    \centering
    \footnotesize
    \begin{tabular}{c|c|c|c|ccc}
    \toprule[0.95pt]
        \multirow{2}{*}{GPH} & \multirow{2}{*}{PPH} & \multirow{2}{*}{PPH\textsuperscript{PVA}} & \multirow{2}{*}{CPH} &  \multicolumn{3}{c}{3D AP\textsubscript{R40} (Mod.)} \\
        ~ & ~ & ~ & ~ & Car & Ped. & Cyc.  \\  \midrule[0.6pt]
        \checkmark & ~ & ~ & ~ & 82.85 & 52.37 & 70.65  \\ 
        ~ &  \checkmark & ~ & ~ & 83.48 & 52.55 & 70.91  \\ 
        ~ & ~ &  \checkmark & ~ & 83.51 & 52.92 & 71.31  \\ 
        ~ & ~ & ~ &  \checkmark & 83.60 & 52.84 & 71.03  \\ 
        ~ &  \checkmark & ~ &  \checkmark & 83.53 & 53.85 & 71.11  \\ 
        ~ & ~ &  \checkmark &  \checkmark & \textbf{83.55} & \textbf{53.91} & \textbf{71.25}  \\ 
        \bottomrule[0.95pt]
    \end{tabular} 
    \vspace{-0.5em}
    \caption{Performance comparison of different implementations of the proposed multi-pooling enhancement module.}   
    \vspace{-1.5em}
    \label{table5}
\end{table}

\section{Conclusion}
\label{sec:conclusion}
In this paper, we proposed a novel two-stage 3D object detector based on the Point-Voxel Attention Fusion Network (PVAFN), which addresses the challenge of 3D object detection by fusing point and voxel representations through contextual information. PVAFN has two main components: first, the proposed point-voxel attention mechanism adaptively fuses the features of points and voxel-BEV representations, capturing rich contextual information to mitigate the limitations of sparse point clouds. Second, in the refinement network stage, the proposed multi-pooling enhancement module not only acquires rich and high-granularity information through a pyramid structure but also focuses on foreground point feature extraction through the clustering pooling method, enabling rapid localization of key geometric features. PVAFN fully leverages the advantages of point and voxel representations, achieving competitive detection performance on the KITTI and Waymo datasets.

\bibliographystyle{IEEEtran} 
\bibliography{ref}

\end{document}